\documentclass[conference]{IEEEtran}

\usepackage{graphicx}
\graphicspath{ {figures/} }
\usepackage{url}
\urldef\spamassassinurl\url{http://csmining.org/index.php/spam-assassin-datasets.html?file=tl_files/Project_Datasets/SpamAssassin%20data/ }
\usepackage[]{algorithm2e}
\usepackage{comment}
\usepackage{mathtools}
\usepackage{subcaption}
\captionsetup[subfigure]{width=0.85\textwidth}
\let\labelindent\relax
\usepackage{enumitem}

\usepackage{fancyhdr}
\usepackage{hyperref}
\fancypagestyle{firstpage}{
  \fancyhf{}
  
  \fancyfoot[L]{{\em This paper was published in the Proceedings of the 2019 IEEE 13th International 
Conference on Semantic Computing (ICSC), pages 287-294, Newport Beach, CA, USA, 2019. \copyright 2019 IEEE\\
Link to article abstract in IEEE Xplore: http://dx.doi.org/10.1109/ICOSC.2019.8665546}}
}
\pagestyle{plain}

\begin{document}

\pagenumbering{gobble}

\title{The Use of Unlabeled Data versus Labeled Data for Stopping Active Learning for Text Classification}

\IEEEoverridecommandlockouts

\author
{
\IEEEauthorblockN{Garrett Beatty\dag \thanks{\dag These students contributed equally to this paper.}}
\IEEEauthorblockA{Department of Computer Science\\
The College of New Jersey\\
Ewing, NJ 08628\\
Email: beattyg2@tcnj.edu}
\and
\IEEEauthorblockN{Ethan Kochis\dag}
\IEEEauthorblockA{Department of Computer Science\\
The College of New Jersey\\
Ewing, NJ 08628\\
Email: kochise1@tcnj.edu}
\and
\IEEEauthorblockN{Michael Bloodgood}
\IEEEauthorblockA{Department of Computer Science\\
The College of New Jersey\\
Ewing, NJ 08628\\
Email: mbloodgood@tcnj.edu}
}

\maketitle

\thispagestyle{firstpage}

\begin{abstract} \label{sec:abstract}
Annotation of training data is the major bottleneck in the creation of text classification systems. Active learning is a commonly used technique to reduce the amount of training data one needs to label.  A crucial aspect of active learning is determining when to stop labeling data. Three potential sources for informing when to stop active learning are an additional labeled set of data, an unlabeled set of data, and the training data that is labeled during the process of active learning. To date, no one has compared and contrasted the advantages and disadvantages of stopping methods based on these three information sources. We find that stopping methods that use unlabeled data are more effective than methods that use labeled data.
\end{abstract}

\section{Introduction} \label{sec:introduction}

The use of active learning to train machine learning models has been used as a way to reduce annotation costs for text and speech processing applications \cite{hantke2017, bloodgood2010ACL, lee2012, mairesse2010, miura2016}.  Active learning has been shown to have a particularly large potential for reducing annotation cost for text classification \cite{lewis1994, tong2001}. Text classification is one of the most important fields in semantic computing and it has been used in many applications \cite{mishler2017ICSC, hoi2006, janik2008, allahyari2014, kanakaraj2015}.

Data annotation is a major bottleneck in developing new text classification systems. Active learning is a method that can be used to reduce this bottleneck whereby the machine actively selects which data to have labeled for training. The careful selection of the data to be labeled enables the machine to learn high performing models from smaller amounts of data than if passive learning were used. The active learning process is shown in Algorithm~\ref{alg:activeLearning}.

An important aspect of the active learning process is the stopping criterion as shown in Algorithm~\ref{alg:activeLearning}.  Stopping methods enable the potential benefits of active learning to be achieved in practice.  Without stopping methods, the active learning process would continue until the entire unlabeled pool has been annotated, which would defeat the purpose of active learning.  Consequently, many stopping methods have been researched to achieve the benefits of active learning in practice \cite{bloodgood2009CoNLL, bloodgood2013CoNLL, schohn2000, zhu2008b, laws2008, vlachos2008, zhu2008a, altschuler2019}.

\begin{algorithm}
\SetAlgoLined
\SetKwInOut{Input}{Input}
\SetKwFor{Loop}{Loop}{}{End Loop}

\Input{\\$U$ = large pool of unlabeled data \\ $L$ = empty pool of labeled data \\
$b$ = batch size}
$L \gets$ select $b$ random examples from $U$ and request their labels\;
$U = U - L$

\Loop{until stopping criterion is met}{
Train model using $L$\;
$batch \gets$ select $b$ examples from $U$ using selection algorithm and request their labels\;
$U = U - batch$\;
$L = L \cup batch$\;
}
\caption{Active Learning Algorithm}
\label{alg:activeLearning}
\end{algorithm}

The purpose of active learning is to reduce the data annotation bottleneck by carefully selecting the data to be labeled. To avoid labeling any additional data, active learning stopping methods have been developed that use only unlabeled data to stop the active learning process. It has been suggested that using labeled data would be a straightforward way to stop the active learning process, but stopping methods using labeled data have not been thoroughly explored because of the extra cost of labeling the data. However, the use of labeled data might make stopping methods so much more effective that the extra cost of the labeled data is worthwhile. To date, investigating whether the advantages of using labeled data outweigh the disadvantages of using labeled data for determining when to stop active learning has not been explored. In this paper, we compare stopping methods using unlabeled data with stopping methods using labeled data to see if the additional cost of labeling the data for the purpose of determining when to stop is worthwhile. We find that not only is the extra labeling cost not worthwhile, but stopping methods using unlabeled data actually perform better than stopping methods using labeled data. 

Section~\ref{sec:methodology} explains our methodology. Section~\ref{sec:relatedWork} discusses related work. Section~\ref{sec:experimentalSetup} provides details about our experimental setup.  Section~\ref{sec:results} presents the results of our experiments and section~\ref{sec:conclusion} concludes.

\section{Methodology} \label{sec:methodology}

\subsection{Stopping Method Information Sources} \label{subsec:stoppingMethodDataSources}

One could classify the information sources that stopping methods use into three categories:

\begin{enumerate}[label={(\roman*)}]
\item unlabeled data, 
\item small labeled data, and
\item training data labeled during the active learning process.
\end{enumerate}

The first category is unlabeled data. Stopping methods that use unlabeled data allow for the full potential of active learning to be realized because a stopping point is found without incurring any additional labeling cost.

The second category of data is a small labeled set.  Following \cite{zhu2008a}, we will refer to this set as a \textit{validation set} in the rest of this paper. Using a validation set to stop the active learning process would appear to be the most direct way to stop the active learning process.  Having a validation set would mean that the performance of the model could be approximated.  However, creating a validation set means annotating examples before the active learning process begins.  This might defeat the purpose of active learning, since examples are being annotated that may not be requested by the selection algorithm throughout the training process.

The third category of data is created during the active learning process: the training data.  Formalized as $L$ in Algorithm~\ref{alg:activeLearning}, this is the data that is labeled in order to train a model.  Since the training data is already labeled, one can use it to determine when to stop active learning without incurring additional labeling cost.

Unlabeled data is a potentially large set of unlabeled examples. Since the examples are unlabeled, the data can be made as large as needed to be as representative of the application space as desired. The validation set does not contain artificial sources of bias and does contain labels, but it has to be relatively small due to the extra labeling cost.  The training data contains labels and can be of moderate size, but it is systematically biased due to how it is selected.  The size of the training set is moderate as it grows over time.  It is not clear which information source, or combination of them, is most effective for stopping active learning.

\subsection{Stopping Methods That Use Unlabeled Data} \label{subsec:stoppingMethodsUnlabeledData}

Several stopping methods for active learning have been researched for the field of text classification.  Schohn and Cohn created a stopping method, which we denote as SC2000, that will stop the active learning process when the model's confidence values of the unlabeled data are outside of the model's margin \cite{schohn2000}. This method can only be used with margin-based learners such as Support Vector Machines (SVMs). Vlachos devised a stopping method, which we denote as V2008, that will stop active learning when the confidence values of the unlabeled data drops consistently for three consecutive models \cite{vlachos2008}.  Laws and Sch{\"u}tze investigated a stopping method, which we denote as LS2008, that will stop active learning when the gradient of model confidence values is less than a user-specified threshold \cite{laws2008}.  The gradient is calculated using the medians of the averages of the confidence values of the selected batches of examples for $k$ iterations of active learning, where $k$ is a user-specified parameter. Zhu, Wang, and Hovy created a stopping method, which we denote as ZWH2008, that uses multiple criteria.  First, it will check if the accuracy on the next batch of training data exceeds a threshold. Then, it will stop active learning when the classifications of the unlabeled pool did not change from the previous model's predictions \cite{zhu2008b}.  Bloodgood and Vijay-Shanker developed the Stabilizing Predictions (SP) stopping method. We denote this method as BV2009.  This method examines the predictions of consecutively trained models on an unlabeled set of data, referred to as a stop set.  The method stops active learning when the agreement of consecutively trained models on the stop set is greater than a user-specified threshold \cite{bloodgood2009CoNLL}. Bloodgood and Grothendieck then improved SP with an added variance check to dynamically adjust the stop set size as needed \cite{bloodgood2013CoNLL}. We denote this method as BG2013.

In \cite{bloodgood2009CoNLL, bloodgood2013CoNLL, wiedemann2018}, and in our results in section~\ref{sec:results}, SP is shown to be a leading stopping method that uses unlabeled data. Therefore, in the rest of this paper, we use the SP stopping method as representative of the state of the art of  stopping methods that use unlabeled data.

\subsection{Stopping Methods That Use Labeled Data} \label{subsec:stoppingMethodsLabeledData}

Several stopping methods have been suggested that use labeled data.  One such method is the Performance Threshold method that will stop the active learning process after the mean of model performance for a user-defined amount of iterations exceeds a user-defined threshold \cite{li2006}.  This method ensures that the model is reaching a performance level that the user deems acceptable.  Another method is the Performance Difference method that will stop the active learning process once the mean of model performance differences for a user-defined amount of iterations is less than a user-defined threshold \cite{bloodgood2009CoNLL, vlachos2008, zhu2008a, li2006}.  This method determines when the performance on the labeled set levels off.  These methods can be used with a validation set and with the training data.  To our knowledge, these methods have never been implemented or tested.

\subsection{Stopping Methods That Use Multiple Data Sources} \label{subsec:stoppingMethodsMultipleData}

Stopping methods that use both unlabeled data and labeled data have not been discussed in previous work.  We combine our labeled data stopping methods with Stabilizing Predictions  \cite{bloodgood2009CoNLL} with the variance check described in \cite{bloodgood2013CoNLL} in four ways:

\begin{enumerate}[label={(\roman*)}]
\item SP $\wedge$ Performance Threshold
\item SP $\wedge$ Performance Difference
\item SP $\vee$ Performance Threshold
\item SP $\vee$ Performance Difference
\end{enumerate}

The SP $\wedge$ Performance Threshold method and the SP $\wedge$ Performance Difference method stop the active learning process when both SP and the labeled data stopping method indicate to stop. The SP $\vee$ Performance Threshold method and the SP $\vee$ Performance Difference method stop the active learning process when either SP or the labeled data stopping method indicate to stop. Using terminology introduced in \cite{bloodgood2009CoNLL}, SP $\wedge$ Labeled Data stopping methods are more {\em conservative} and the stopping points are guaranteed to be at least as late as $max$(SP stopping point, labeled data stopping point). On the other hand, SP $\vee$ Labeled Data stopping methods are more {\em aggressive} and stop at least as early as $min$(SP stopping point, labeled data stopping point).

\section{Related Work} \label{sec:relatedWork}

\subsection{Using Unlabeled Data for Stopping}

Past work using unlabeled data for stopping active learning was discussed in section~\ref{subsec:stoppingMethodsUnlabeledData}.

\subsection{Using Labeled Data for Stopping}

A validation set, or a small labeled set, is one way of stopping the active learning process \cite{schohn2000}. Labeling data that might not be used in the training process, however, defeats the purpose of active learning \cite{schohn2000}.  Determining the size of the validation set is an open question \cite{zhu2008a}.  If the validation set is too small, it might not be representative of what can be learned, resulting in skewed stopping points \cite{vlachos2008, zhu2008a}.  However, making a larger validation set would increase the cost, defeating the purpose of active learning \cite{zhu2008a}.  We investigate different validation set sizes in section~\ref{subsec:validationSetSize}. Although stopping using a validation set has been discussed as a possibility, to our knowledge, stopping methods using labeled data have never been implemented or tested.  We examine the performance of stopping methods that use a validation set in section~\ref{subsec:validVSUnlabeled}.  Also, although we don't want our separate held-out test set to have any overlap with training examples, it is less clear whether the advantages of allowing the training set to overlap with the validation set outweigh the disadvantages. We explore this in section~\ref{subsec:validationSetReplacement}.

Using cross-validation on the training set has been discussed as an information source for stopping methods.  Schohn and Cohn \cite{schohn2000} stated that the time needed to re-train an SVM model would make this information source impractical to use.  They also stated that the distribution created by the training set might not be representative of the test set distribution \cite{schohn2000}. This means that data collected from the cross-validation on the training set could be skewed in relation to the data collected from a test set.  It is known that actively sampled data can be quite skewed from randomly sampled data \cite{bloodgood2009NAACL}.  However, using data labeled for training has the advantage of being able to use relatively large amounts of labeled data without incurring any additional cost.  To our knowledge, previous work has not examined using the training data to stop active learning.  We examine the performance of stopping methods that use cross-validation on the training data in section~\ref{subsec:traincvVSUnlabeled}.

\subsection{Other Related Work}

Small labeled sets have also been used in other areas of active learning.  A small labeled set can be used to estimate the ratio of negative to positive examples in an entire corpus to build a cost-weighted SVM \cite{bloodgood2009NAACL}. Neural networks can stop training by using the performance score on a small labeled set \cite{kotsiantis2007, prechelt1998, domhan2015}. Finally, a small labeled set can be used to build a biased SVM when no negative examples are present in the training set \cite{liu2003}.  None of these works considered using small labeled sets to stop the active learning process, which we experiment with in section~\ref{subsec:validVSUnlabeled}.

\section{Experimental Setup} \label{sec:experimentalSetup}

We use the 20NewsGroups dataset\footnote{Downloaded the ``bydate" version from \url{http://qwone.com/~jason/20Newsgroups/} on July 13, 2017.  This version does not include duplicate posts and is sorted by date into train and test sets.}, the Reuters dataset\footnote{Downloaded from \url{http://www.daviddlewis.com/resources/testcollections/reuters21578/} on July 13, 2017.}, the WebKB dataset\footnote{Downloaded from \url{http://www.cs.cmu.edu/afs/cs.cmu.edu/project/theo-20/www/data/} on July 13, 2017.}, and the spamassassin corpus\footnote{Downloaded the latest versions of the 5 distinct sets from \spamassassinurl \ on July 13, 2017.} for our experiments. For the Reuters dataset, we use the ten largest categories from the Reuters-21578 Distribution 1.0 ModApte split, as in \cite{joachims1998} and \cite{dumais1998}. Consistent with previous work, we report the results for the four largest categories of the WebKB dataset \cite{mccallum1998}, \cite{zhu2008a}, \cite{zhu2008b}.  Averages for the 20NewsGroups and Reuters datasets were taken across the categories.  Averages for the categories of SpamAssassin and WebKB were taken over a 10-fold cross-validation.  We use a Support Vector Machine as our classifier and use the closest-to-hyperplane selection algorithm \cite{schohn2000, tong2001, campbell2000}. This selection algorithm was recently found to have better performance than other selection algorithms \cite{bloodgood2018ICSC}.  We use a batch size that is equivalent to 0.5\% of the initial unlabeled pool for each dataset and keep adding this amount of new examples for each iteration of active learning.  For text classification, we use binary features for every word that occurs more than three times and remove stop words that appear in the \textit{Long Stopword List} from https://www.ranks.nl/stopwords.

\subsection{Validation Set} \label{subsec:validationSet}

We build the validation set by randomly selecting examples from the unlabeled pool.  When using a validation set, two important questions arise: 
\begin{enumerate}[label={(\roman*)}]
\item How big should the validation set be?
\item Should examples from the validation set be allowed to be selected for training during active learning?
\end{enumerate}
In section~\ref{sec:results} we present results of experiments investigating these questions.

\subsection{Training Data Cross Validation} \label{subsec:tenFoldCVTrain}

As mentioned in section~\ref{sec:methodology}, training data itself could be used by stopping methods.  In order to do this, we use 10-fold cross-validation (CV) on the training data, as shown in Algorithm~\ref{alg:trainTenFold}. 

\begin{algorithm}
\SetAlgoLined
\SetKwInOut{Input}{Input}
\SetKwFor{Loop}{Loop}{}{End Loop}

\Input{\\$U$ = large pool of unlabeled data \\ $L$ = empty pool of labeled data \\
$b$ = batch size \\ $p$ = empty array of performance scores \\ $p\_avg$ = empty array of performance score averages}
$L \gets$ select $b$ random examples from $U$ and request their labels\;
$U = U - L$

\Loop{until stopping criterion is met}{
$S \gets$ partition $L$ into 10 sets\;
\For{$i\leftarrow 1$ \KwTo $10$}{
$model_i \leftarrow$ Train model using $S - S[i]$\;
$p[i]$ = Test $model_i$ using $S[i]$\;
}
$p\_avg \leftarrow$Average $p[1\ldots10]$\;
$batch \gets$ select $b$ examples from $U$ using selection algorithm and request their labels\;
$U = U - batch$\;
$L = L \cup batch$\;
}
\caption{Active Learning Algorithm Using 10-fold CV on $L$}
\label{alg:trainTenFold}
\end{algorithm}

\subsection{Stopping Method Parameters}

The Performance Difference method uses $\epsilon$ as its threshold of F-Measure difference between the active learning iterations.  A larger value of $\epsilon$ would cause the method to be more aggressive, as it would stop when the performance is still increasing at a faster rate.  A smaller value of $\epsilon$ would cause the method to be more conservative, as it would only stop when performance changes have become smaller.  By default, we use an $\epsilon$ value of 0.005: half of a percentage point of F-Measure.  Half of a percentage point of F-Measure was chosen as a default value for $\epsilon$ since learning will be relatively stable, while still allowing for some fluctuations due to noise and random events.

The Performance Threshold method uses $\tau$ as its threshold value. This value is representative of the performance level of the model the user wants to achieve.  A larger value of $\tau$ would lead to a more conservative method.  A smaller value of $\tau$ will cause the method to be more aggressive.  By default, we set $\tau$ to 0.8, or 80\% F-Measure.  In many cases, a model that has a performance level of 80\% F-Measure is considered reasonable. Setting $\tau$ is more difficult and dataset-dependent than setting $\epsilon$ because the level of performance that is acceptable depends heavily on the task and dataset whereas the level of $\epsilon$ that indicates a leveling off in performance is not so heavily dependent on the task and dataset. 

Both the Performance Difference and the Performance Threshold method look back $w$ iterations of active learning to determine if the models' performance on the validation set has leveled off or has sustained a user-defined level of performance for $w$ iterations.  A relatively small value of $w$ would mean that the models' performance does not have to be stable or above a certain value for many iterations of active learning. If $w$ is too small, the method becomes more aggressive.  A larger value of $w$ would mean that the performance needs to be stable or above a certain value over more iterations, which would help avoid the risk of stopping too early.  However, using a larger $w$ means one would need more labeled data for the increased number of iterations, causing the method to become more conservative.  Following previous work, we set $w$ to three \cite{bloodgood2009CoNLL, vlachos2008}.  As \cite{beatty2018ICSC} advised, if a relatively large batch size is used, a smaller value for $w$ should be used in order to mitigate the degradation in stopping method performance caused when using larger batch sizes. 
\section{Results} \label{sec:results}

\subsection{Unlabeled Stopping Methods} \label{subsec:unlabeledResults}

Table~\ref{table:unlabeled} shows the performance of unlabeled data stopping methods. SP, one of the most widely applicable and easy-to-implement methods, has leading performance, consistent with past findings \cite{bloodgood2009CoNLL, bloodgood2013CoNLL, wiedemann2018}. Accordingly, we use SP as representative of state of the art unlabeled data stopping methods in the rest of our experiments.

\begin{table*}
\resizebox{\textwidth}{!}{
\begin{tabular}{| l | c | c | c | c | c | c |}
\hline
{Datasets} & {SP (BV2009/BG2013)} & {SC 2000} & {V 2008} & {LS 2008} & {ZWH 2008} & {ALL}\\
\hline
{20NewsGroups} & {823} & \textbf{1915} & {748} & \textbf{513} & {877} & {11314} \\\cline{2-7} {(20-cat AVG)} & {73.36} & {74.34} & \textbf{47.24} & {67.09} & {73.64} & {74.59} \\
\hline
{Reuters} & {691} & \textbf{1267} & {2286} & {628} & {739} & {9655} \\\cline{2-7} {(10-cat AVG)} & {77.94} & {78.12} & \textbf{58.59} & {71.60} & {78.18} & {77.70} \\
\hline
{SpamAssassin-spam} & {294} & \textbf{847} & \textbf{5441} & {1292} & \textbf{378} & {5441} \\\cline{2-7} {(10-fold AVG)} & {98.10} & \textbf{98.78} & \textbf{98.91} & {96.34} & {98.47} & {98.91} \\
\hline
{WebKB-course} & {669} & \textbf{1332} & {2314} & \textbf{370} & \textbf{810} & {7445} \\\cline{2-7} {(10-fold AVG)} & {84.96} & {86.12} & \textbf{68.35} & \textbf{75.49} & {85.83} & {83.44} \\
\hline
{WebKB-faculty} & {728} & \textbf{1306} & {614} & \textbf{325} & \textbf{950} & {7445} \\\cline{2-7} {(10-fold AVG)} & {86.29} & {87.22} & \textbf{68.52} & \textbf{78.95} & {86.86} & {85.38} \\
\hline
{WebKB-project} & {806} & \textbf{1335} & {1366} & \textbf{229} & {858} & {7445} \\\cline{2-7} {(10-fold AVG)} & {66.29} & {67.53} & \textbf{53.24} & \textbf{43.28} & {66.52} & {65.57} \\
\hline
{WebKB-student} & {1039} & \textbf{2009} & \textbf{4937} & \textbf{262} & \textbf{1428} & {7445} \\\cline{2-7} {(10-fold AVG)} & {83.31} & {84.59} & {79.16} & \textbf{72.43} & {84.38} & {83.81} \\
\hline
{Average} & {722} & \textbf{1430} & \textbf{2529} & {517} & \textbf{863} & {8027} \\\cline{2-7} {(Macro AVG)} & {81.46} & {82.39} & \textbf{67.72} & \textbf{72.17} & {81.98} & {81.35} \\
\hline
\end{tabular}
}
\caption{Unlabeled data methods for stopping AL. For each dataset, the average number of annotations at the automatically determined
stopping points and the average F-measure at the automatically determined stopping points are displayed. \textbf{Bold} entries
are statistically significantly different than SP (and non-bold entries are not). The Average row is simply an unweighted
macro-average over all the datasets. The final column (labeled ``All") represents standard fully supervised passive
learning with the entire set of training data.} 
\label{table:unlabeled}
\end{table*}

\subsection{Effect of Allowing Validation Set Examples to be Selected for Training} \label{subsec:validationSetReplacement}

There is a potential validation set performance estimation bias\footnote{Note there is no test set performance estimation bias since the test set is a completely held-out separate set of data with no overlap with any other data.} when validation set examples are allowed to be selected for training. We examine the impact of allowing validation set examples to be selected for training on the performance metric F-Measure using a validation set size of 500.

\begin{figure}
    \centering
    \includegraphics[width=0.5\textwidth]{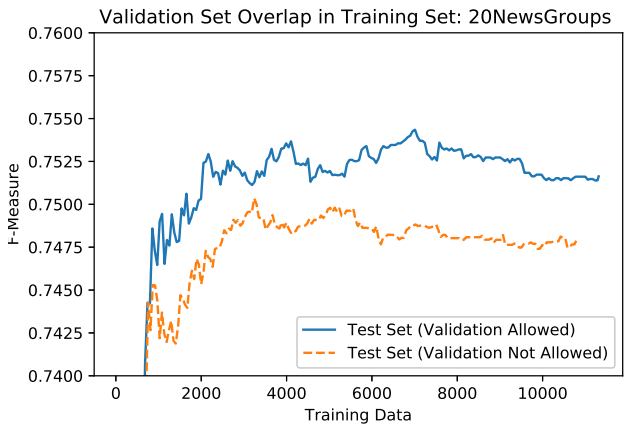}
    \caption{Test set F-Measure when validation set examples are allowed to be selected for training versus when they are not allowed to be selected for training using a validation set size of 500. There is only one test set, however, two lines are shown because two separate sets of models were trained (one set that is allowed to select validation set examples for training and one set that is not). The dotted gold line stops exactly 500 (size of the validation set) examples earlier than the solid blue line because examples from the validation set were not available to train that set of models.}
    \label{fig:20NewsGroupszoomed}
    \vspace{-.25cm}
\end{figure}

There are two main benefits when validation set examples are allowed to be selected for training. The first benefit is that the overall test set performance is higher, as seen in Figure~\ref{fig:20NewsGroupszoomed}. The reason for this performance increase is because high value examples that were in the validation set were allowed to be used for training. If validation set examples were not allowed to be selected for training, the model's learning efficiency may be hurt because the high value examples can not be used to improve the model. The second benefit is that when a training example is selected from the validation set, it can be used without any extra labeling cost. 

\begin{figure*}[t!]
    \centering
    \begin{subfigure}[t]{0.5\textwidth}
        \centering
         \includegraphics[width=0.95\textwidth]{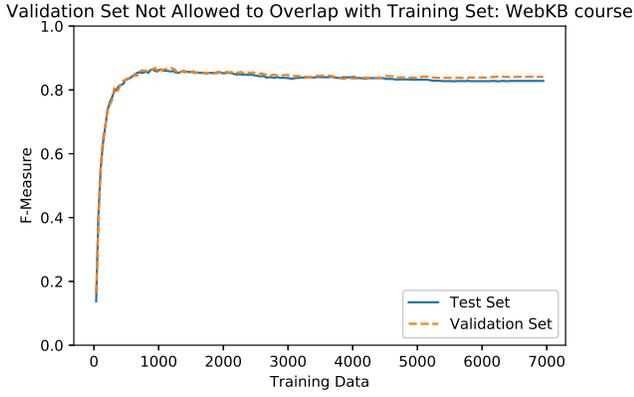}
         \caption{Test set and validation set F-Measure when validation set examples are not allowed to be selected for training using a validation set size of 500.}
	 \label{fig:WebKBCourseStoppingValNotAllowed}
    \end{subfigure}%
    ~ 
    \begin{subfigure}[t]{0.5\textwidth}
        \centering
        \includegraphics[width=0.9\textwidth]{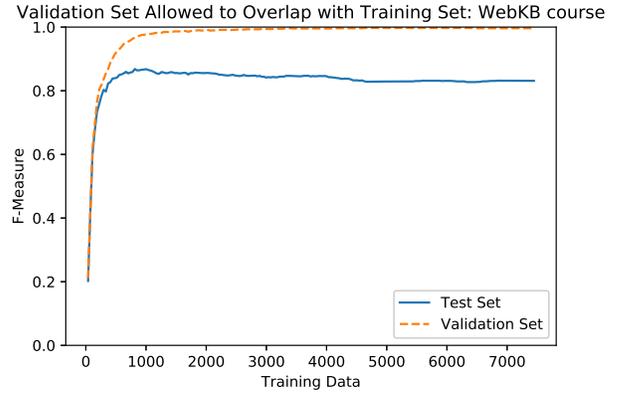}
    	\caption{Test set and validation set F-Measure when validation set examples are allowed to be selected for training using a validation set size of 500.}
	\label{fig:WebKBCourseStoppingValAllowed}
    \end{subfigure}
    \caption{Validation Set performance estimation curves when examples from the validation set are allowed to be selected as training data and when examples from the validation set are not allowed to be selected as training data.}
    \label{fig:WebKBCourseStoppingValAllowedOrNot}
\end{figure*}

As mentioned before, the other option is to not allow validation set examples to be selected for training. The main benefit of this approach is that the validation set estimate of performance will be a better approximation of test set performance, as seen in Figure~\ref{fig:WebKBCourseStoppingValAllowedOrNot}. The reason that it more closely approximates the test set performance than the first approach is because there is no performance estimation bias from examples in the training data also being in the validation set. 

Note, however, that the validation set performance curves in Figure~\ref{fig:WebKBCourseStoppingValAllowedOrNot} qualitatively have the same shape when examples are allowed in the training data as when they're not allowed. The Performance Difference method can use this behavior effectively to determine when to stop.  The Performance Threshold method would not be able to use this behavior, but this does not matter because the Performance Threshold method does not perform well in our experiments anyway (e.g., see Table~\ref{table:validationSet}) since it is tough to set the threshold value of $\tau$. Therefore, the benefits of allowing validation set examples to be selected for training outweigh the drawbacks, and we allow examples from the validation set to be selected for training. Note that in all cases all final performance values in all of our experiments are computed using a completely held-out separate test set.

\subsection{Size of Validation Set} \label{subsec:validationSetSize}

\begin{figure}
    \centering
    \includegraphics[width=0.5\textwidth]{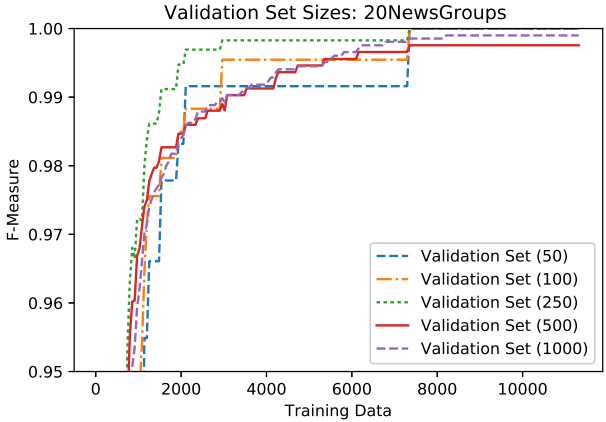}
    \caption{Validation set F-Measure validation set sizes: 50, 100, 250, 500, 1000. Smaller validation set sizes are shown to be more erratic.}
    \label{fig:20NewsGroupsstoppingvalwithreplacement}
\end{figure}

\begin{figure}
    \centering
    \includegraphics[width=0.5\textwidth]{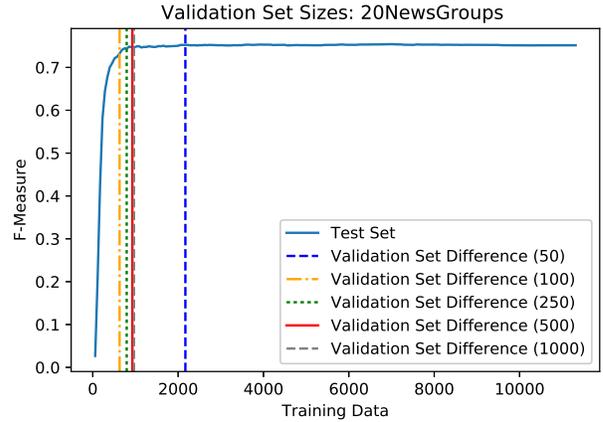}
    \caption{Test set F-Measure for validation set stopping methods for multiple validation set sizes: 50, 100, 250, 500, 1000. Stopping methods (from left to right): 100, 250, 500, 1000, 50.}
    \label{fig:20NewsGroupsstoppingvalsizes}
\end{figure}

The size of the validation set should be large enough to be representative, but small enough to be cost-efficient. To test the effect that size has on validation set stopping methods, we computed validation set performance using validation sets with sizes of 50, 100, 250, 500, and 1000.  In Figure~\ref{fig:20NewsGroupsstoppingvalwithreplacement}, we can see that performance estimates using validation set sizes of 50, 100, and 250 are erratic compared to performance estimates using sizes of 500 and 1000. The effect that this erratic behavior has on stopping methods can be seen in Figure~\ref{fig:20NewsGroupsstoppingvalsizes}, where stopping methods that use smaller validation set sizes perform poorly. From Figure~\ref{fig:20NewsGroupsstoppingvalwithreplacement} one can see that increasing the size of the validation set to be larger than 500 costs more labels, but does not improve performance estimates.  In the rest of our experiments, we use a validation set size of 500.

\subsection{Validation Set and Unlabeled Data Stopping Methods} \label{subsec:validVSUnlabeled}

Table~\ref{table:validationSet} shows the performance of validation set stopping methods and unlabeled data stopping methods. In Table~\ref{table:validationSet} we can see that validation set stopping methods tend to have worse performance than unlabeled data stopping methods. We can also see that SP $\wedge$ validation set stopping methods stop at a later iteration than unlabeled data stopping methods. This is expected because as mentioned in section~\ref{subsec:stoppingMethodsMultipleData}, SP $\wedge$ validation set stopping methods are more conservative and are guaranteed to stop later than or at the same point of SP. We can also see that SP $\vee$ validation set stopping methods stop earlier than or at about the same iteration than unlabeled data stopping methods. Once again, this is expected because SP $\vee$ validation set stopping methods are more aggressive and are guaranteed to stop earlier than or at the same point as SP. 

\begin{table*}
\resizebox{\textwidth}{!}{
\begin{tabular}{| l | c | c | c | c | c | c | c |}
\hline
{Datasets} & {SP (BV2009/BG2013)} & {Threshold} & {Difference} & {SP $\wedge$ Threshold} & {SP $\wedge$ Difference} & {SP $\vee$ Threshold} & {SP $\vee$ Difference}\\
\hline
{20NewsGroups} & {846} & \textbf{461} & {929} & {846} & {957} & \textbf{461} & {817} \\\cline{2-8} {(20-cat AVG)} & {74.92} & {70.29} & {74.94} & {74.92} & {75.16} & {70.29} & {74.70} \\
\hline
{Reuters} & {662} & \textbf{355} & {628} & {662} & {758} & \textbf{355} & {590} \\\cline{2-8} {(10-cat AVG)} & {79.47} & {76.91} & {78.75} & {79.47} & {79.21} & {76.91} & {78.66} \\
\hline
{SpamAssassin-spam} & {291} & \textbf{86} & {270} & {291} & {299} & \textbf{86} & {264} \\\cline{2-8} {(10-fold AVG)} & {98.70} & \textbf{91.03} & {98.26} & {98.70} & {98.63} & \textbf{91.03} & {98.33} \\
\hline
{WebKB-course} & {703} & \textbf{273} & {680} & {703} & \textbf{780} & \textbf{273} & {625} \\\cline{2-8} {(10-fold AVG)} & {86.27} & \textbf{79.89} & {84.85} & {86.27} & {86.16} & \textbf{79.89} & {84.96} \\
\hline
{WebKB-faculty} & {736} & \textbf{266} & {703} & {736} & {802} & \textbf{266} & {677} \\\cline{2-8} {(10-fold AVG)} & {86.42} & \textbf{82.59} & {86.08} & {86.42} & {86.73} & \textbf{82.59} & {85.94} \\
\hline
{WebKB-project} & {828} & \textbf{562} & {788} & {828} & \textbf{917} & \textbf{562} & {736} \\\cline{2-8} {(10-fold AVG)} & {67.76} & {64.43} & {66.53} & {67.76} & {67.05} & {64.43} & {66.89} \\
\hline
{WebKB-student} & {1047} & \textbf{373} & \textbf{817} & {1047} & {1102} & \textbf{373} & {817} \\\cline{2-8} {(10-fold AVG)} & {84.55} & \textbf{79.18} & \textbf{82.08} & {84.55} & {84.60} & \textbf{79.18} & {82.08} \\
\hline
{Average} & {730} & \textbf{339} & {688} & {730} & {802} & \textbf{339} & {647} \\\cline{2-8} {(Macro AVG)} & {82.58} & {77.76} & {81.64} & {82.58} & {82.50} & {77.76} & {81.65} \\
\hline
\end{tabular}
}
\caption{SP versus Validation Set Stopping Methods.  For each dataset, the average number of annotations at the automatically determined
stopping points and the average F-measure at the automatically determined stopping points are displayed. \textbf{Bold} entries
are statistically significantly different than SP (and non-bold entries are not). The Average row is simply an unweighted
macro-average over all the datasets. Performance Threshold has been renamed to ``Threshold" and Performance Difference has been renamed to ``Difference" to fit the table on the page.} 
\label{table:validationSet}
\end{table*}

Overall, unlabeled data stopping methods perform similarly or better than both validation set stopping methods and stopping methods that combine both the validation set and unlabeled data.

\subsection{Training Set CV and Unlabeled Data Stopping Methods} \label{subsec:traincvVSUnlabeled}

Table~\ref{table:trainCV} shows the performance of training set CV stopping methods and unlabeled data stopping methods. In Table~\ref{table:trainCV} we can see that training set CV stopping methods tend to have worse performance than unlabeled data stopping methods. SP $\wedge$ Training Set CV stopping methods stop more conservatively than unlabeled data stopping methods. On the other hand, SP $\vee$ Training Set CV stopping methods stop more aggressively than unlabeled data stopping methods. 

\begin{table*}
\resizebox{\textwidth}{!}{
\begin{tabular}{| l | c | c | c | c | c | c | c |}
\hline
{Datasets} & {SP (BV2009/BG2013)} & {Threshold} & {Difference} & {SP $\wedge$ Threshold} & {SP $\wedge$ Difference} & {SP $\vee$ Threshold} & {SP $\vee$ Difference}\\
\hline
{20NewsGroups} & {823} & {864} & \textbf{1459} & \textbf{2694} & \textbf{1615} & \textbf{316} & {803} \\\cline{2-8} {(20-cat AVG)} & {73.36} & \textbf{60.95} & {73.96} & {74.39} & {74.29} & \textbf{60.66} & {73.27} \\
\hline
{Reuters} & {691} & \textbf{187} & {859} & {734} & {964} & \textbf{187} & {600} \\\cline{2-8} {(10-cat AVG)} & {77.94} & \textbf{58.01} & {77.22} & {77.97} & {78.68} & \textbf{58.01} & {76.68} \\
\hline
{SpamAssassin-spam} & {294} & \textbf{89} & \textbf{753} & {313} & \textbf{753} & \textbf{89} & {294} \\\cline{2-8} {(10-fold AVG)} & {98.10} & \textbf{91.58} & \textbf{98.78} & {98.15} & \textbf{98.78} & \textbf{91.58} & {98.10} \\
\hline
{WebKB-course} & {669} & \textbf{303} & \textbf{1139} & \textbf{1568} & \textbf{1139} & \textbf{199} & {669} \\\cline{2-8} {(10-fold AVG)} & {84.96} & \textbf{70.61} & {85.93} & {85.65} & {85.93} & \textbf{70.37} & {84.96} \\
\hline
{WebKB-faculty} & {728} & \textbf{299} & \textbf{1354} & \textbf{1572} & \textbf{1417} & \textbf{214} & {710} \\\cline{2-8} {(10-fold AVG)} & {86.29} & \textbf{73.68} & {86.70} & {87.45} & {86.85} & \textbf{73.56} & {86.17} \\
\hline
{WebKB-project} & {806} & \textbf{6000} & \textbf{1509} & \textbf{7445} & \textbf{1509} & {673} & {806} \\\cline{2-8} {(10-fold AVG)} & {66.29} & {61.42} & {67.19} & {65.57} & {67.19} & {61.18} & {66.29} \\
\hline
{WebKB-student} & {1039} & {1957} & {1361} & \textbf{3167} & \textbf{1698} & \textbf{669} & {950} \\\cline{2-8} {(10-fold AVG)} & {83.31} & {78.14} & {83.60} & {84.11} & {84.51} & \textbf{77.76} & {83.03} \\
\hline
{Average} & {722} & {1386} & \textbf{1205} & \textbf{2499} & \textbf{1299} & \textbf{335} & {690} \\\cline{2-8} {(Macro AVG)} & {81.46} & \textbf{70.63} & {81.91} & {81.90} & {82.32} & \textbf{70.44} & {81.22} \\
\hline
\end{tabular}
}
\caption{SP versus Training Set CV Stopping Methods.  For each dataset, the average number of annotations at the automatically determined
stopping points and the average F-measure at the automatically determined stopping points are displayed. \textbf{Bold} entries
are statistically significantly different than SP (and non-bold entries are not). The Average row is simply an unweighted
macro-average over all the datasets. Performance Threshold has been renamed to ``Threshold" and Performance Difference has been renamed to ``Difference" to fit the table on the page.} 
\label{table:trainCV}
\end{table*}

Overall, unlabeled data stopping methods perform similarly or better than both training set CV stopping methods and methods that combine both the training set CV and unlabeled data.

\section{Conclusion} \label{sec:conclusion}
Active learning has the potential to significantly reduce annotation costs for text classification. One of the main considerations in the active learning process is when to stop the iterative process of asking for more labeled data. Previous work has researched stopping methods that use unlabeled data. Using labeled data in the form of a small labeled validation set or using cross-validation on the training set as information sources for stopping methods was considered but not tested in previous work, leaving an open question of whether labeled data stopping methods would perform sufficiently better than unlabeled data stopping methods in order to justify any additional expenses associated with gathering the labeled data to inform the stopping method. We performed an investigation of stopping methods based on labeled data, unlabeled data, and combinations. We found that unlabeled data stopping methods are convincingly better than labeled data stopping methods. In our experiments, not only is the extra labeling cost not worthwhile, but stopping methods using unlabeled data perform better than stopping methods using labeled data.
\section*{Acknowledgment} \label{acknowledgment}

This work was supported in part by The College of New Jersey Support of Scholarly Activities (SOSA) program and by The College of New Jersey Mentored Undergraduate Summer Experience (MUSE) program. The authors acknowledge use of the ELSA high performance computing cluster at The College of New Jersey for conducting the research reported in this paper. This cluster is funded by the National Science Foundation under grant number OAC-1828163.

\bibliographystyle{IEEEtran}
\bibliography{ms}

\end{document}